\documentclass{AISB2008}
\usepackage{times}
\usepackage{graphicx}
\usepackage{latexsym}

\begin{document}

\title{A multilayer reactive system for robots interacting with children with autism}

\author{Pablo G\'{o}mez Esteban \institute{Vrije Universiteit Brussel,
Belgium, email: pablo.gomez.esteban@vub.ac.be}, Hoang-Long Cao$^1$ , Albert De Beir$^1$  , \\ Greet Van de Perre$^1$ ,  Dirk Lefeber$^1$ \and Bram Vanderborght$^1$   }

\maketitle
\bibliographystyle{AISB2008}

\begin{abstract}
There is a lack of autonomy on traditional Robot-Assisted Therapy systems interacting with children with autism. To overcome this limitation a supervised autonomous robot controller is being built. In this paper we present a multilayer reactive system within such controller. The goal of this Reactive system is to allow the robot to appropriately react to the child's behavior creating the illusion of being alive.

\end{abstract}

\section{INTRODUCTION} \label{sec:introduction}

Robot-Assisted Therapy (RAT) is widely used, particularly with children with special needs, see \cite{feil2008robot} and \cite{simut2012trends} as examples, reducing the workload of the therapy and therefor its cost. While the benefits of using RAT are undisputed, current approaches  \cite{Huijnen2016} typically constrain themselves to the Wizard of Oz (WOZ) paradigm \cite{landauer1987psychology} \cite{wilson1988rapid}, where the robot is remotely controlled by a human operator, usually the therapist. According to \cite{scassellati2012robots}, for a long-term use the WOZ framework is not a sustainable technique. Robots in RAT are required to become more autonomous in order to reduce cost and time within the therapeutic interventions, see \cite{Thill2012}.

\begin{figure}[thb]
\begin{center}
\includegraphics[scale=0.33]{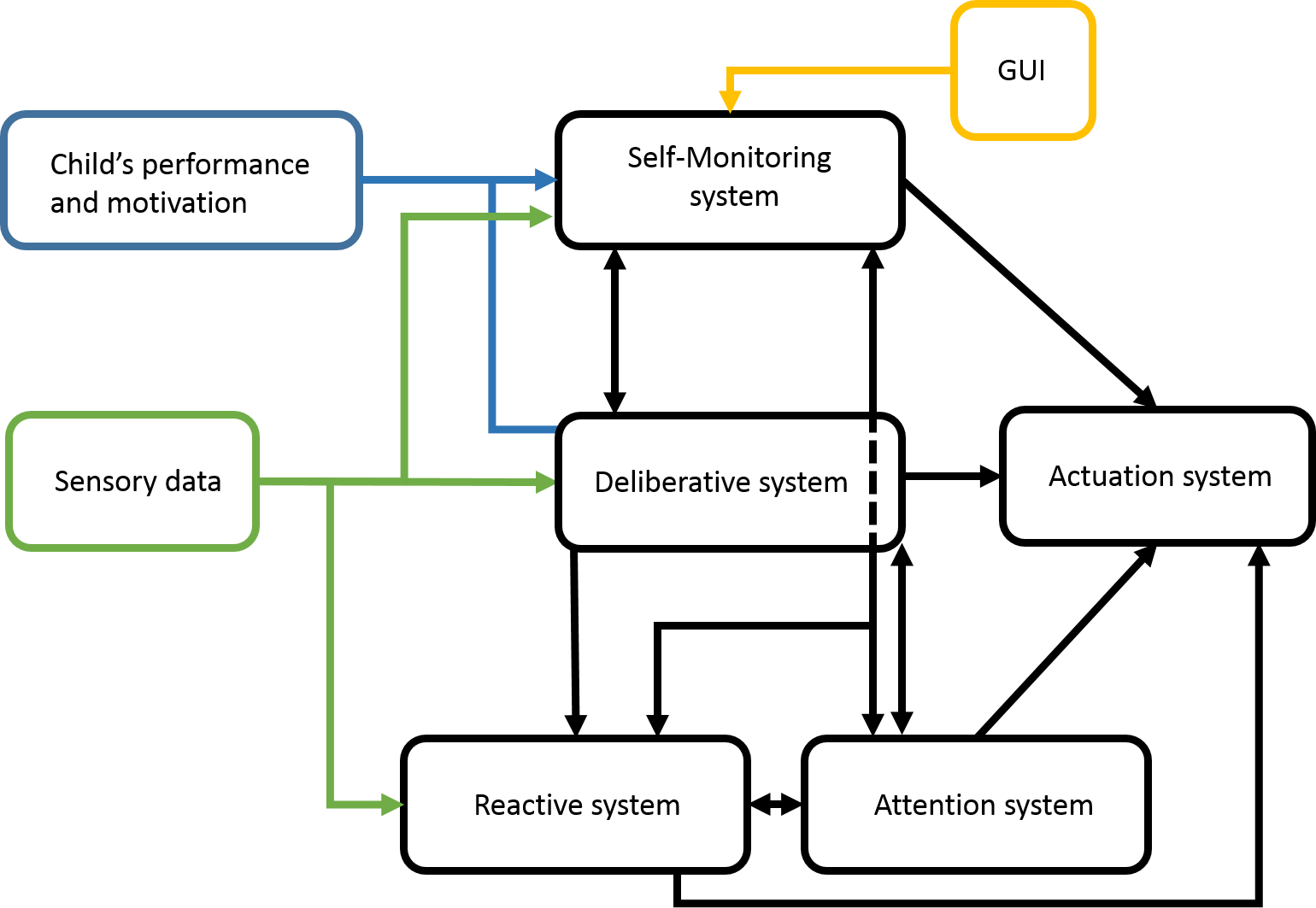}
\end{center}
\caption{Project DREAM's cognitive architecture is composed of several systems (in black) and complemented by an assessment of the child performance and motivation (in blue) and sensory data (in green). Therapist can control the cognitive architecture through a GUI (in yellow). Arrows show flow of information between the systems.}
\label{fig:DREAM}                                                                       
\end{figure}

Under such circumstances the DREAM project (Development of Robot-Enhanced therapy for children with AutisM spectrum disorders) was conceived. This project is concerned, among other research challenges, with the development of an autonomous controller. Despise full autonomy is currently unrealistic, a ``supervised autonomy", where the operator gives the robot certain goals and the robot autonomously works towards achieving them, is certainly feasible. This controller is composed of a number of systems: Reactive, Attention, Deliberative, Self-monitoring and Actuation, see  Figure \ref{fig:DREAM}, and complemented by sensory data and  a module to assess the performance and motivation of the child. The focus of this paper is on the Reactive system.

The Reactive system is constituted of the lowest-level processes. In natural systems, these processes are genetically determined and not typically sensitive to learning. State information, coming from the sensory inputs, is immediately acted upon with appropriate motor outputs. The Reactive system, while absent in many robot systems, is essential in social robots, see \cite{lazzeri2013towards}. It creates the illusion of the robot being alive \cite{saldien2014motion}, and acts as a catalyst for acceptance and bonding between the young user and the robot. It ensures that the robot can handle the real time challenges of its environment appropriately taking care of small motions, appropriate eye blinking, whole body motion during gesturing and head motion, recovering from falls, and appropriately reacting to affective displays by young users. The behaviors will be configurable by the therapist as it might not be desirable for some children to have the robot display a full gamut of reactive responses (for example, a negative reaction when being pushed).

This paper is structured as follows.  In Section \ref{sec:reactive} a high level description of the system is provided. Through subsections \ref{sec:falling} to \ref{sec:blinking} the different layers composing the Reactive system are detailed. Finally, some future work is provided in Section \ref{sec:discussion}.

%
%

%

%
%

\section{A MULTILAYER REACTIVE SYSTEM}\label{sec:reactive}

\begin{figure}[thb]
\begin{center}
\includegraphics[scale=0.27]{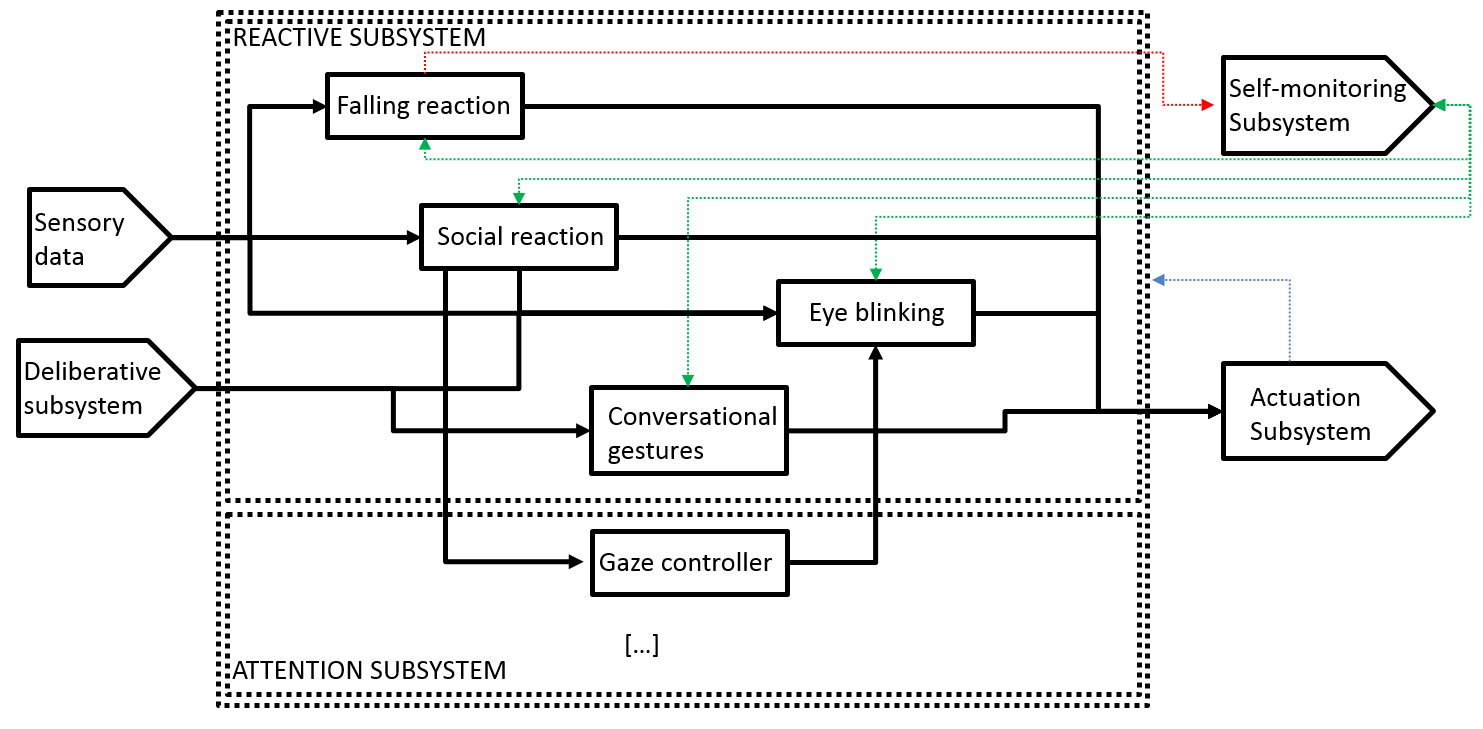}
\end{center}
\caption{High level description of the Reactive system. It receives inputs from the Deliberative system but mainly from the sensory data, and produces its output towards the Actuation system. The Self-monitoring system allows the therapist, through a GUI (see Figure \ref{fig:DREAM}), to switch on/off the functionality of each layer (green arrows). The falling reaction layer might send a signal to interrupt all running behaviors (red arrow). The Actuation system provides feedback about the execution of the motor commands (blue arrow). The remaining arrows show information flow between the layers.}
\label{fig:reactive}                                                                       
\end{figure}

A general high level description of the Reactive system is shown in Figure \ref{fig:reactive}. This describes how, given the sensory information the robot reacts to the current situation. Such information is processed by different layers producing each own outputs towards the Actuation system which will combine them all, according to predefined priorities, to produce the final outcome of the cognitive architecture.

The Reactive system is composed of a number of layers: 
\begin{itemize}
\item The falling reaction triggers a damage avoidance posture when falling. At that moment it interrupts all the running behaviors. Once the robot is back at its feet, it takes care of restoring the intervention behavior.
\item The social reaction purpose is to appropriately react to social displays of the children and to provide small motions and face/sound tracking features that will give the impression of the robot being alive.
\item The eye blinking layer provides a variable blinking rate that complements other gestures and behaviors.
\item Conversational gestures complement the speech acts with body gestures.
\end{itemize}

The therapist might consider that one or more of these layers are not appropriate for being used with a certain child, for such reason, their functionality, which are detailed in the following sections, can be switched on and off when needed through the Self-monitoring system.

\begin{figure}[thb!]
\begin{center}
\includegraphics[scale=0.44]{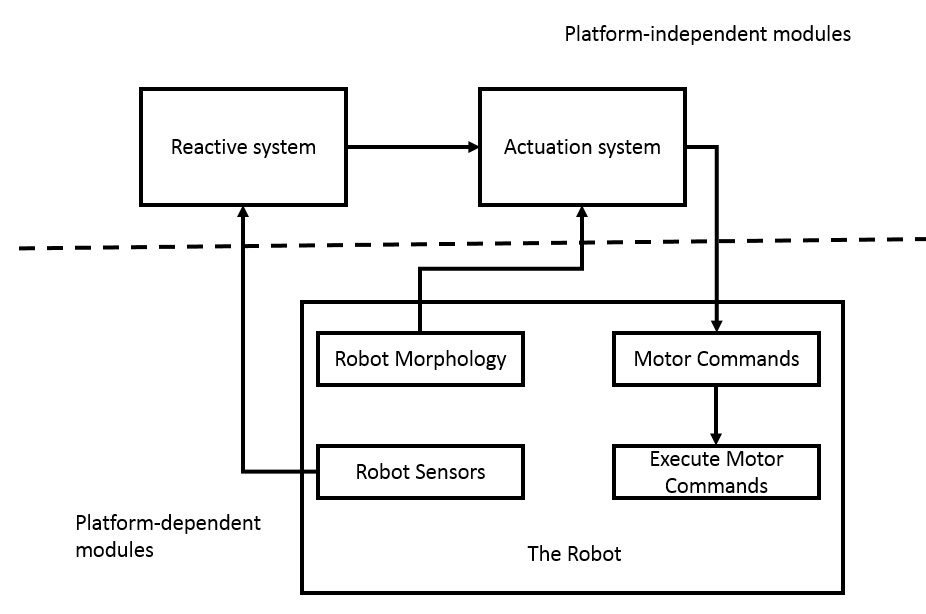}
\end{center}
\caption{The Reactive system provides different outputs to the Actuation system. This system has access to the morphology and hence the degrees of freedom of the robot. }
\label{fig:reactiveSchema}                                                                       
\end{figure}

One of the main contributions of this system is that it can be easily implemented in different robots due to its platform-independence flavor. The Actuation system is responsible of generating the appropriate motor commands depending on robot morphology. This system, see Figure \ref{fig:reactiveSchema}, has access to the degrees of freedom of the robot and generates the corresponding motor commands, see \cite{van2015development} for further details.

\subsection{Falling Reaction}\label{sec:falling}

Within social interaction with children it may happen that robots lose their balance and have to recover it or even they may fall down. These robotic platforms are expensive so that in case they fall, minimizing the hardware damage would be a priority. According to the intervention protocol we aim to use within DREAM project, the robot will be seated in front of the child, so that a fall is lowly probably to occur. Nevertheless, this module needs to be implemented to face such hypothetical situations.

\begin{figure}[thb!]
\begin{center}
\includegraphics[scale=0.43]{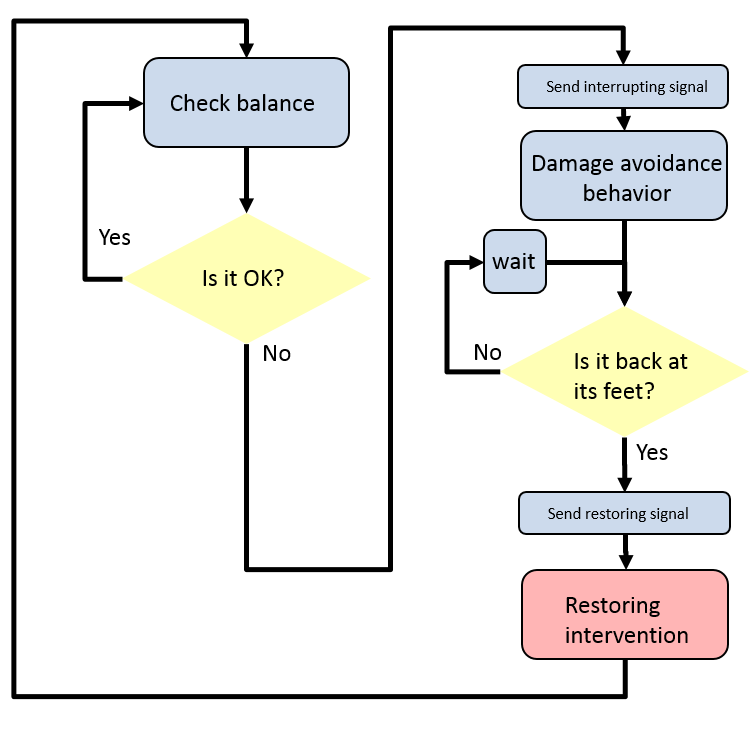}
\end{center}
\caption{The module is periodically checking the balance of the robot. In case of a fall, a signal will be sent to interrupt any other running behavior, and a damage avoidance behavior will be triggered. Finally, back at its feet, the module will send a signal to restore the intervention.}
\label{fig:fallingReaction}                                                                       
\end{figure}

The Falling Reaction module, see Figure \ref{fig:fallingReaction}, will be periodically checking the balance of the robot using the sensory information available. Changes in the balance may end up in a fall. In such case, a signal will be sent to interrupt any other running behavior, and a damage avoidance behavior that fits the situation will be triggered, see \cite{fujiwara2002} for a case of minimizing damage to a humanoid robot, and \cite{yun2012} for a case of a NAO robot that modifies its falling trajectory to avoid causing injuries in people in front of it. These behaviors might be highly dependent on the morphology of the robot. Reducing the stiffness of the joints will avoid any mechanical problem independently of its morphology. 
 Additionally, the robot should include some speech acts to reduce the impact of such dramatic situation for the kid as saying that it has been a little bit clumsy or that it is tired today.

Finally, back at its feet, the robot may apologize in order to engage the child back to the intervention and it will send a signal to restore the system functionality.

\subsection{Social Reaction}\label{sec:social}

In social situations multiple verbal and non-verbal interactive encounters may occur. The child may behave friendly with the robot affectively touching it or may feel unfavorable to it and eventually hit it. These situations may be very conflicting as a special care must be paid with the potential audience of this system. If it would be the case of a regular social robot, for such both situations the robot may appropriately react, but under these circumstances, the reaction will be simplified to facial expressions and speech acts, always under the supervision of the therapist who might consider that such social reaction is not therapeutically appropriate for a specific child. Moreover, in order to reach an effective social interaction, emulating certain degree of empathy towards the social partner plays a key role in patient-centered therapy \cite{tapus2007}, i.e. if the child is expressing an emotion, the robot should be aware of that and react accordingly expressing a compatible emotion. In those cases in which there is no social interaction, this module will randomize among a set of small motions to recreate a life-like behavior such as a breathing motion, gaze-shifts  or sound and/or face tracking. The purpose of this module is to provide the appropriate social behavior in order to give the impression of the robot being socially alive. 

This module receives as input the sensory information where it is specified the child's social and affective state i.e. whether she/he is expressing an emotion or is performing a physical behavior (such as touching the robot unexpectedly). For each of these behaviors there should be a set of facial expressions and speech acts available to choose among them. Ideally it should randomize among them in order to look less predictable. 

\subsection{Conversational gestures}\label{sec:conversational}

Exhibiting co-verbal gestures would make the robot appear more expressive  and intelligible which will help to build social rapport with their users \cite{meena2012}. 

Co-verbal gestures are defined as the spontaneous gestures that accompany human speech, and have been shown to be an integral part of human-human interactive communications  \cite{mcneill1992}. There exist evidences that co-verbal gestures have a number of positive effects performed by robots \cite{salem2013err}\cite{heerink2010relating}.

\begin{figure}[thb!]
\begin{center}
\includegraphics[scale=0.63]{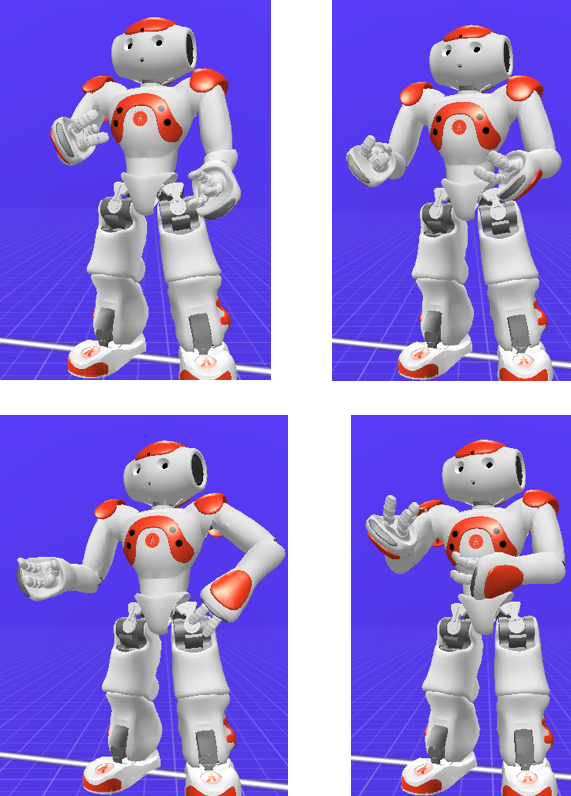}
\end{center}
\caption{Set of conversational gestures belonging to Kendon's Open Hand Supine family related to offering and giving.}
\label{fig:gestures}                                                                       
\end{figure}

We adopted gestures from \cite{csapo2012multimodal} where authors use Kendon's Open Hand Supine (``palm up") family of gestures which are related to offering and giving, see Figure \ref{fig:gestures} where our set of conversational gestures implemented in the Nao robot are shown. As explained in Section \ref{sec:introduction}, we did not consider any negative gesture, as those belonging to the Open Hand Prone (``palm down") family, as it might be not appropriate for this audience.

For the purposes of DREAM project we don't aim at building a highly sophisticated conversational agent as \cite{meena2012} or \cite{bremner2009} but to complement speech acts with conversational gestures, that the robot can randomly perform while speaking trying to improve the acceptability of the robot during the social interaction. For that reasons, we include a set of conversational gestures along with the rules to trigger them.

\subsection{Eye Blinking}\label{sec:blinking}

The acceptability of the robot can be further increased if the robot mimics the human blinking behavior. Simulating blinking behavior requires a human-level blinking model that should be derived from real data of human.

Several works have been done  concerning the dependencies of human eye blinking behavior on different physiological and psychological factors. Ford et al. \cite{ford2013} proposed the ``blink model" for HRI, which integrates blinking as a function of communicative behaviors. Doughty \cite{doughty2001} described in his work three distinct blinking patterns during reading, during conversation and while idly looking at nothing specific. Lee et al. \cite{lee2002} proposed a model of animated eye gaze that integrates blinking as depending on eye movements constituting gaze direction.

\begin{figure}[thb!]
\begin{center}
\includegraphics[scale=0.42]{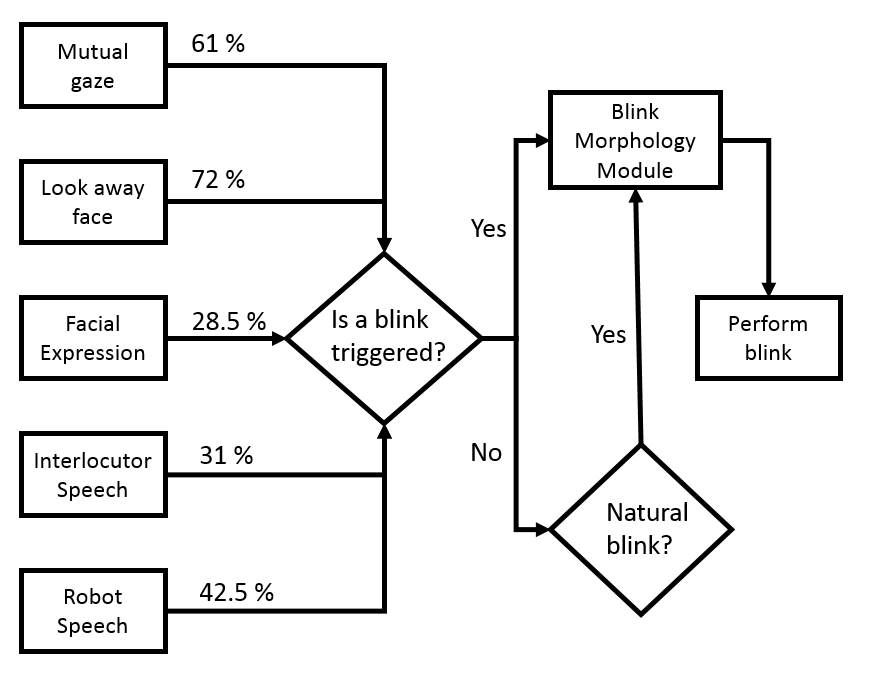}
\end{center}
\caption{Adapted blinking model. When a communicative facial behavior occurs there is a probability of triggering a blink behavior. Such probabilities come from \cite{ford2013}.}
\label{fig:blinkingBehavior}                                                                       
\end{figure}

Given the amount of studies made to model human blinking behavior we don't need to do our own but to use that one that best fits our requirements. Within the context in which DREAM will be applied, we need to recreate a blinking behavior mainly focused on the communicative behaviors and gaze shifts. For such reason, we have simplified and adapted Ford et al.'s model to our needs, see Figure \ref{fig:blinkingBehavior}, defining a model which considers multiple communicative facial behaviors. For each of them there is a probability to blink. Moreover there is a passive behavior which simulates a natural, or non-interactive, blinking mechanism (for cleaning or humidifying the eye) that can be activated when no other blinking behavior has been triggered. To perform the blinking motion there is a blink morphology module which defines, based on statistics, if the blink is simple or multiple, full or half, its duration, etc. 

\section{CONCLUSIONS AND FURUTE WORK}\label{sec:discussion}

In this paper we present a multilayer Reactive system within an autonomous robot controller. The goal of such system is to allow the robot to appropriately react to the child's behavior creating the illusion of being alive. For such purpose it is composed of several layers that can be switched on and off by the therapist depending the needs of the intervention: the falling reaction layer is aimed to prevent and manage falls; the social reaction one to appropriately react to social displays; another one to provide a blinking behavior to complement gestures; and, finally, some conversational gestures to complement speech acts.

This controller has a platform-independent flavor which allows it to be implemented in multiple robotic platforms without spending too much effort on it. Some test on Nao, Pepper and Romeo are about to be made. Also, studies on the acceptability of this system are under development.

Lip synchronization in robotics looks for matching lip movements with the audio generated by the robot. Several works use synchronization algorithms based directly on the use of audio phonemes to determine the levels of mouth aperture \cite{oh2010} \cite{hara1997}. These approaches require additional information such as dictionaries of phonemes. Currently all the robots available to this research group  to implement this system have no mouth. As future work we aim to implement a basic lip synchronization method like \cite{oh2010} in the second version of the huggable robot Probo \cite{debeir2016evolutionary} which is currently under development.

\section*{ACKNOWLEDGEMENTS}
The work leading to these results has received funding from the European Commission 7th Framework Program as a part of the project DREAM grant no.
611391.

\bibliography{DREAM}

\end{document}